\documentclass[APA,LATO1COL]{WileyNJD-v2}

\articletype{Original Article}%

\usepackage{graphicx}
\usepackage{hyperref}
\usepackage{booktabs}
\usepackage{numprint}

\hypersetup{hidelinks}

\usepackage{soul}
\usepackage{color}
\DeclareRobustCommand{\hlA}[1]{{\sethlcolor{green}\hl{#1}}}
\DeclareRobustCommand{\ctB}[1]{{\color{blue}#1}}
\renewcommand\hlA[1]{#1} 

\begin{document}

\title{Joint one-sided synthetic unpaired image translation and segmentation for colorectal cancer prevention\protect\thanks{This project has received funding from the European Union’s Horizon 2020 research and innovation programme under the Marie Skłodowska-Curie grant agreement No 765140. This publication has emanated from research supported by Science Foundation Ireland (SFI) under Grant Number SFI/12/RC/2289\_P2, co-funded by the European Regional Development Fund.}}

\author[1,2]{Enric Moreu}

\author[1,2]{Eric Arazo}

\author[1,2]{Kevin McGuinness}

\author[1,2]{Noel E. O'Connor}

\authormark{Enric Moreu \textsc{et al}}

\address[1]{\orgname{Insight SFI Centre for Data Analytics}, \orgaddress{\country{Ireland}}}

\address[2]{\orgname{Dublin City University}, \orgaddress{\country{Ireland}}}


\corres{Enric Moreu \linebreak \email{enric.moreu2@mail.dcu.ie}}


\abstract[Summary]{
Deep learning has shown excellent performance in analysing medical images. However, datasets are difficult to obtain due privacy issues, standardization problems, and lack of annotations. We address these problems by producing realistic synthetic images using a combination of 3D technologies and generative adversarial networks. 
We propose CUT-seg, a joint training where a segmentation model and a generative model are jointly trained to produce realistic images while learning to segment polyps. We take advantage of recent one-sided translation models because they use significantly less memory, allowing us to add a segmentation model in the training loop. CUT-seg performs better, is computationally less expensive, and requires less real images than other memory-intensive image translation approaches that require two stage training.
Promising results are achieved on five real polyp segmentation datasets using only one real image and zero real annotations.
As a part of this study we release Synth-Colon, an entirely synthetic dataset that includes 20000 realistic colon images and additional details about depth and 3D geometry:\\ \url{https://enric1994.github.io/synth-colon}

}


\jnlcitation{\cname{%
\author{E. Moreu}, 
\author{E. Arazo}, 
\author{N. O'Connor}, and
\author{K. McGuinness}}, (\cyear{2022}), 
\ctitle{Joint one-sided synthetic unpaired image translation and segmentation for colorectal cancer prevention}, \cjournal{Expert Systems}, \cvol{e13137}. \url{https://doi.org/10.1111/exsy.13137}}

\maketitle


\section{Introduction}\label{sec1}
Colorectal cancer is one of the most commonly diagnosed cancer types. It can be treated with an early intervention, which consists of detecting and removing polyps in the colon. The accuracy of the procedure strongly depends on the medical professional's experience and hand-eye coordination during the procedure, which can last up to 60 minutes. Computer vision can provide real-time support for doctors to ensure a reliable examination by double-checking all the tissues during the colonoscopy.

\begin{figure}
\centering
\includegraphics[width=90px]{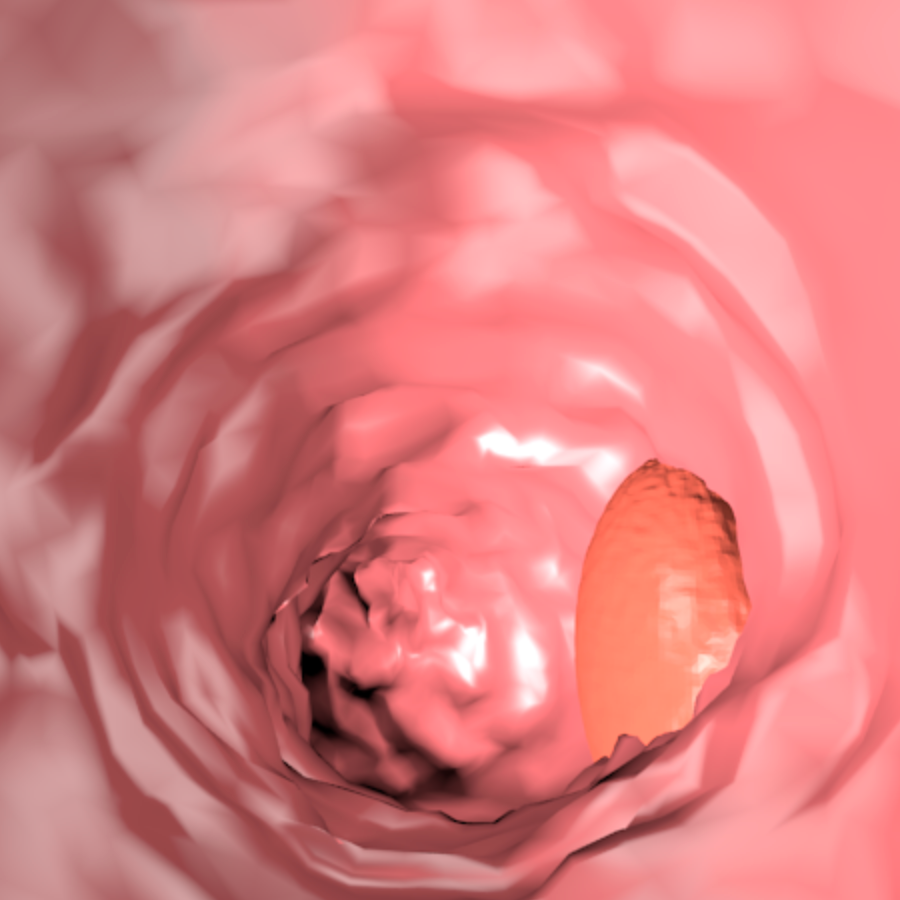}
\includegraphics[width=90px]{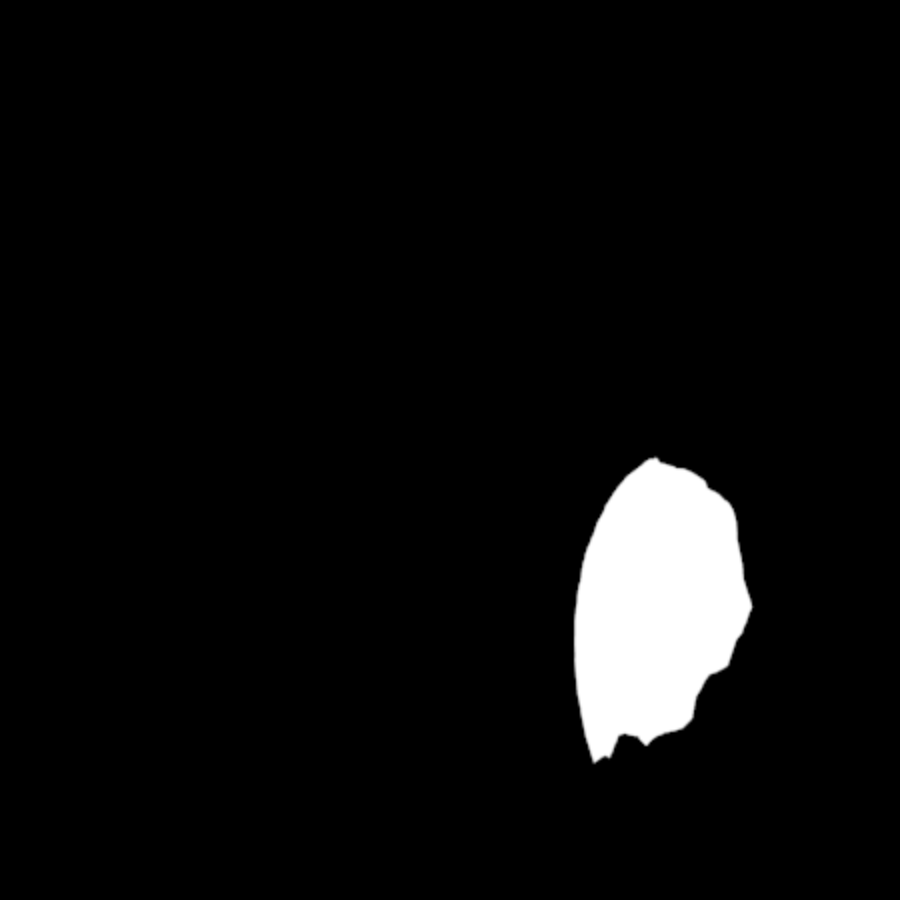}
\includegraphics[width=90px]{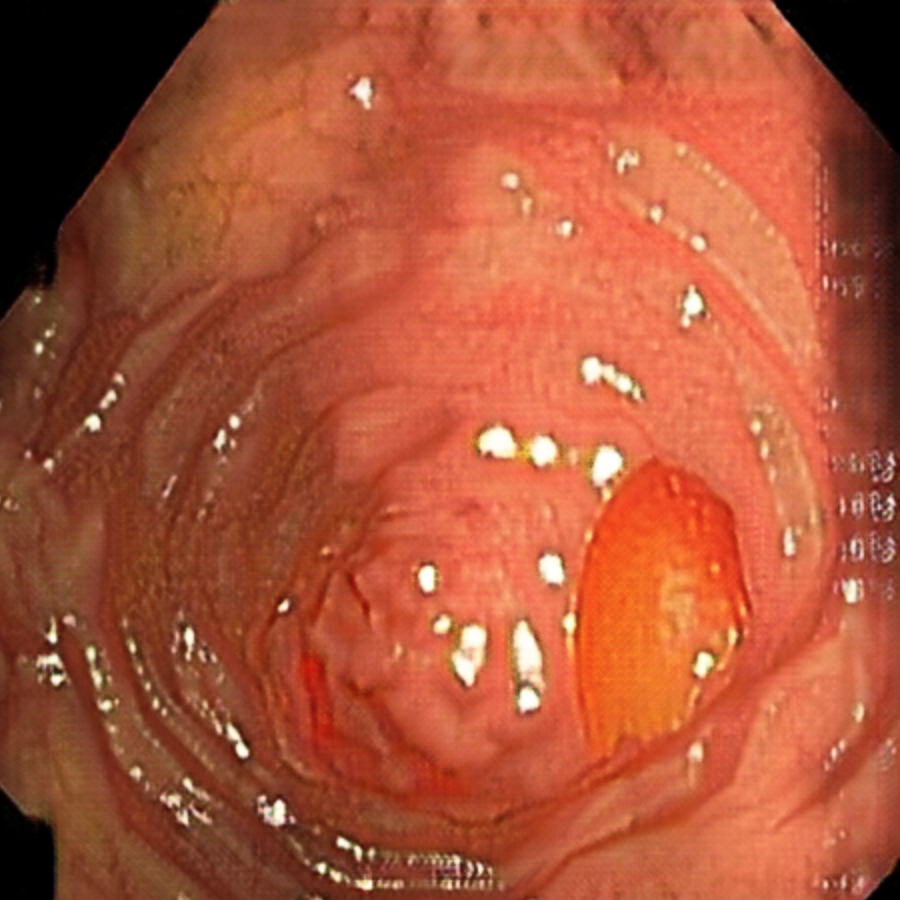}
\includegraphics[width=90px]{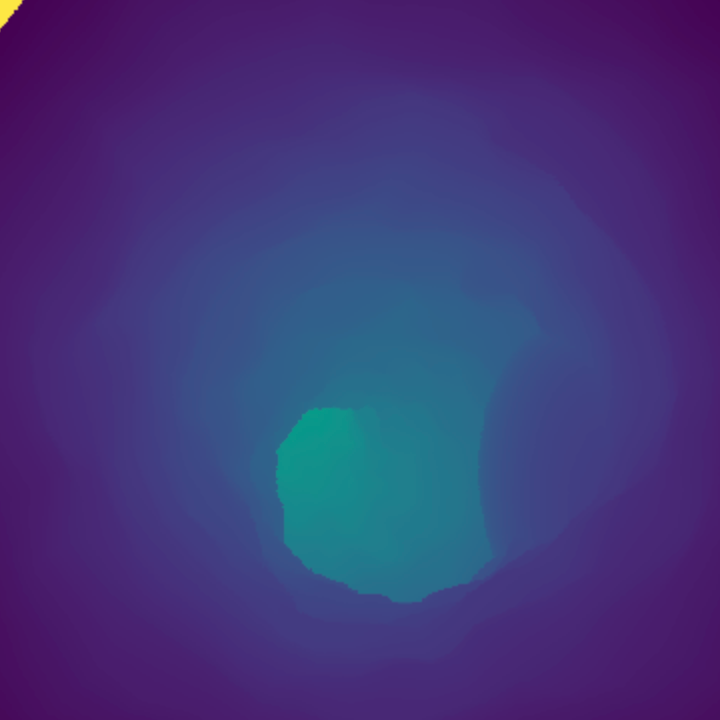}
\includegraphics[width=90px]{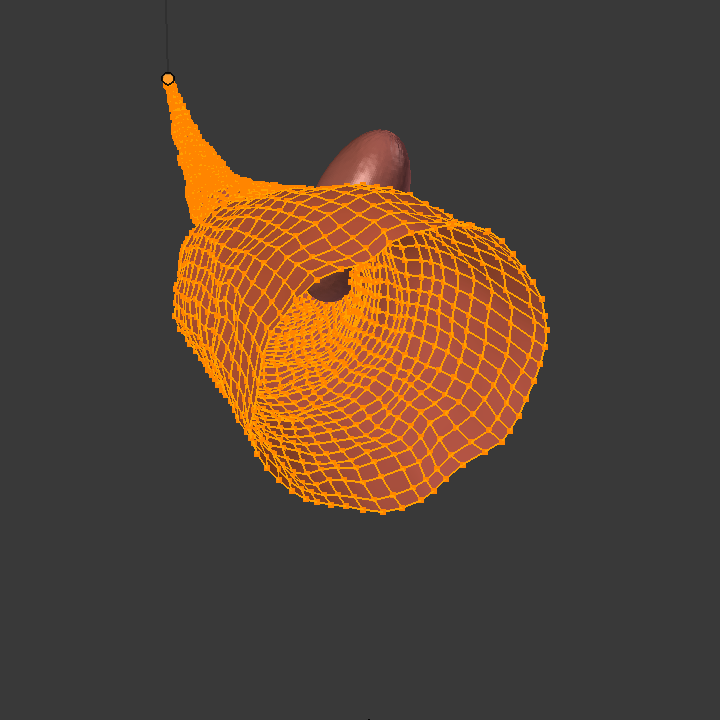}
\caption{Synth-Colon dataset samples: synthetic image, annotation, realistic image, depth map, and 3D mesh (from left to right).} \label{dataset}
\end{figure}

The data obtained during a colonoscopy is accompanied by a set of issues that prevent creating datasets for computer vision applications. \hlA{Firstly}, due to the sensitive nature of the content and data privacy issues, it cannot be used without the consent of the patients.  \hlA{Secondly}, there \hlA{is} a wide range of cameras and lights used to perform colonoscopies. Every device has its own focal length, aperture, and resolution. There are no large datasets with standardized parameters.  Finally, polyp segmentation datasets are expensive because they depend on the annotations of highly qualified professionals.

We propose a method for polyp segmentation that does not require human annotations, by combining 3D rendering and generative adversarial networks.  This paper is an extension of our work ``Synthetic data for unsupervised polyp segmentation''~\citep{2021_AICS_synthColon} and introduces an improved approach and additional insights on the utilization of synthetic data for polyp segmentation. 
\hlA{Firstly}, we produce Synth-Colon, an artificial dataset of colons and polyps generated using 3D rendering. Annotations of the location of the polyps are automatically generated by the 3D engine. \hlA{Secondly}, we combine Synth-Colon with real images from colonoscopies to train a polyp segmentation model. This model learns to translate synthetic images into real ones and then perform the segmentation. In the initial baseline, the image translation part was carried out by a CycleGAN model~\citep{CycleGAN2017} and it was trained independently from the segmentation one. In this extension we unify these two stages and propose CUT-seg, which jointly trains a HarDNeT-based~\citep{chao2019hardnet} segmentation model and a contrastive unpaired translation (CUT)~\citep{park2020cut} image translation model. It transforms synthetic images to the real domain while, at the same time, learning to segment the polyps. Moreover, even with a single real image, CUT-seg performs better than the CycleGAN-based baseline. By comparing CUT-seg with the CycleGAN-based model, we evaluate their ability to translate synthetic images to the real domain and perform polyp segmentation.

The contributions of this paper are as follows:
\begin{itemize}

\item
To the best of our knowledge, we are the first to successfully train a polyp segmentation model with zero manual annotations and reduce the performance gap between methods that use manually annotated labels and those that do not. Additionally, we obtain competitive results when training with a single image from the real world.

\item 
We propose a novel architecture that jointly trains a one-sided image translation model and a segmentation model.

\item 
We release Synth-Colon (see Figure \ref{dataset}), the largest synthetic dataset for polyp segmentation including additional data such as depth and 3D mesh.

\end{itemize}

The remainder of the paper is structured as follows: Section 2 reviews relevant work; Section 3 explains our method; Section 4 presents the Synth-Colon dataset; Section 5 describes our experiments; and Section 6 concludes the paper.

\section{Related work}\label{sec2}
This section briefly reviews relevant works on polyp segmentation, synthetic data, and unpaired image translation.

\subsection{Polyp segmentation}

\begin{figure}
\centering
\includegraphics[width=110px]{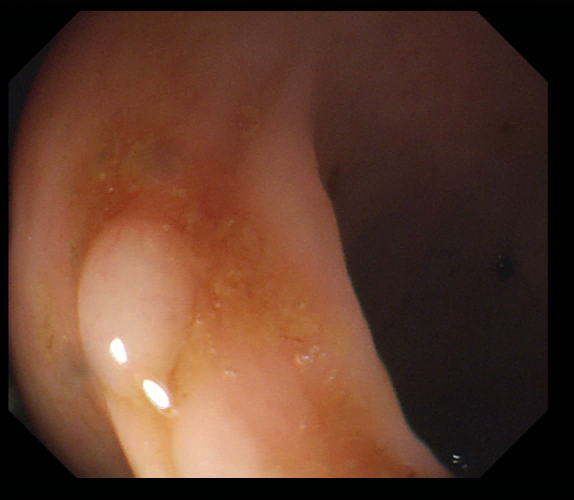}
\includegraphics[width=110px]{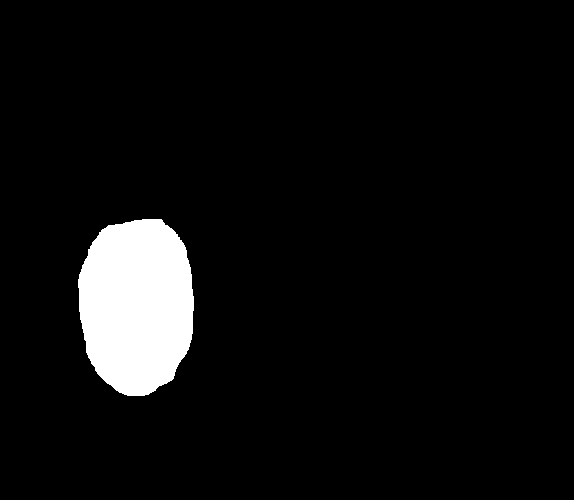}
\includegraphics[width=110px]{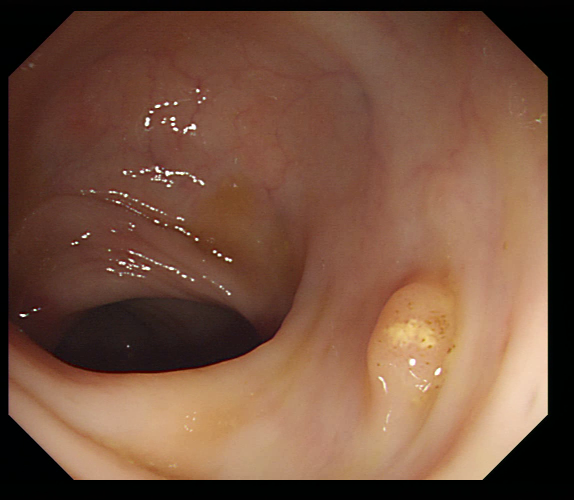}
\includegraphics[width=110px]{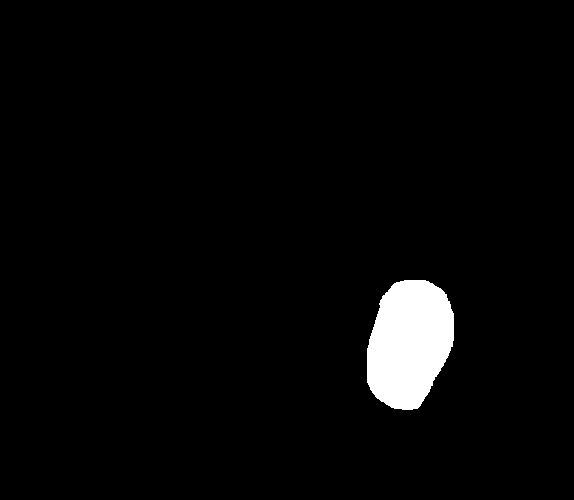}
\caption{Samples of real images from CVC-ColonDB with the corresponding annotation made by \hlA{medical professionals} indicating the location of cancerous polyps.} \label{real_samples}
\end{figure}

\hlA{There are three types of potentially malignant polyps, also known as neoplastic polyps: tubular, villous and villotubular} ~\cite{shinya1979morphology}. \hlA{This categorization refers to the shape and configuration of the polyps. At the same time, polyps are categorized as sessile or pedunculated depending on the degree of attachment to the walls of the intestines.} \hlA{Hence, detecting polyps can be challenging because they present various textures and shapes.}
\hlA{Early work on neoplastic polyp segmentation used ellipse fitting techniques based on shape} ~\citep{hwang2007polyp}. However, some colorectal polyps can be small (5mm) and are not detected by these techniques. In addition, polyp texture is easily confused with other tissues in the colon as can be seen in Figure \ref{real_samples}.

With the rise of convolutional neural networks (CNNs) \citep{lecun2015deep} the challenge of the texture and shape of the polyps was largely solved and the performance was substantially increased. Several works have applied deep convolutional networks to the polyp segmentation problem. \citet{brandao2017fully} proposed to use a fully convolutional neural network based on the VGG \citep{simonyan2014very} architecture to identify and segment polyps. Unfortunately, the small datasets available and the large number of parameters make these large networks prone to overfitting. \citet{zhou2018unet++} used an encoder-decoder network with dense skip pathways between layers that prevented the vanishing gradient problem of VGG networks. They also significantly reduced the number of parameters, reducing the amount of overfitting. More recently, \citet{chao2019hardnet} reduced the number of shortcut connections in the network to speed-up inference time, a critical issue when performing real-time colonoscopies in high-resolution. They focused on reducing the memory traffic to access intermediate features, reducing the latency. Finally, \citep{huang2021hardnetmseg} improved the performance and inference time by combining HarDNet~\citep{chao2019hardnet} with a cascaded partial decoder~\citep{wu2019cascaded} that discards larger resolution features of shallower layers to reduce latency.

\subsection{Synthetic data}

The limitation of using a large CNN is that it often requires large amounts of annotated data. \hlA{This problem is particularly acute in medical imaging due to privacy issues, standardization, and the lack of professional annotators} ~\citep{hardy2021intraprocedural}. Table~\ref{datasets} shows the size and the resolution of the datasets used to train and evaluate existing polyp segmentation models. Polyp datasets are small compared to other computer vision datasets with millions of images. This lack of large datasets for polyp segmentation can be addressed by generating synthetic data~\citep{thambawita2021singan, rottshaham2019singan}. For instance, \citet{thambawita2021singan} used a generative adversarial network (GAN) to produce new colonoscopy images and annotations. They added a fourth channel to SinGAN \citep{rottshaham2019singan} to generate annotations that are consistent with the colon image. They then used style transfer to improve the realism of the textures. Their results are excellent considering the small quantity of real images and professional annotations that are used. \citet{gao2020adaptive} used a CycleGAN to translate colonoscopy images to polyp masks. In their work, the generator learns how to segment polyps by trying to fool a discriminator.

\begin{table}[t]
\centering
\caption{Size and resolutions of polyp segmentation datasets containing real images.}\label{datasets}
\begin{tabular}{lcc}
\toprule
Dataset &  \#Images & Resolution\\
\midrule
CVC-T \citep{vazquez2017benchmark} & 912 & 574 $\times$ 500\\
CVC-ClinicDB \citep{bernal2015wm} &  612 & 384 $\times$ 288\\
CVC-ColonDB \citep{tajbakhsh2015automated} & 380 & 574 $\times$ 500\\
ETIS-LaribPolypDB \citep{silva2014toward} & 196 & 1225 $\times$ 966\\
Kvasir \citep{jha2020kvasir} & 1000 & Variable\\
\bottomrule
\end{tabular}
\end{table}

\begin{figure}
\centering
\includegraphics[width=150px]{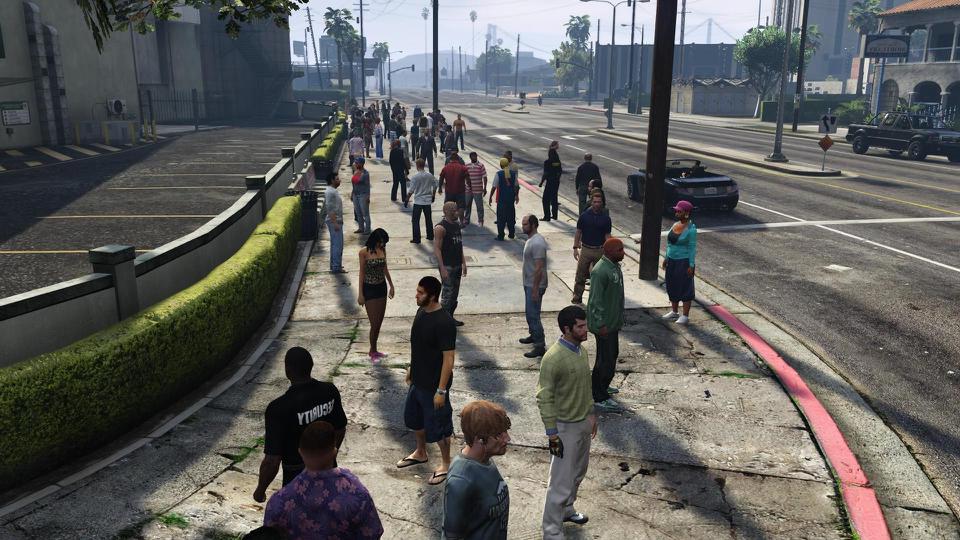}
\includegraphics[width=150px]{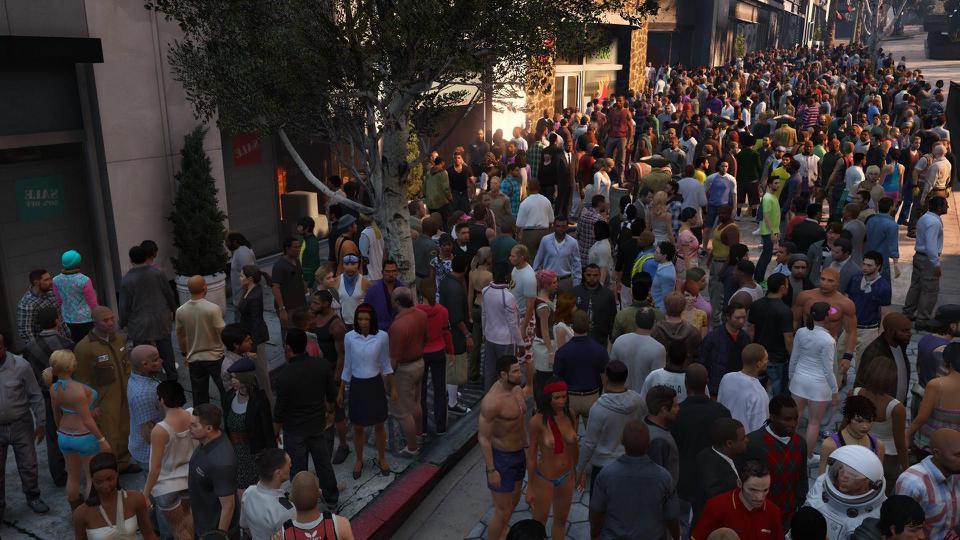}
\caption{Crowd counting synthetic dataset sample images generated by the GTA V videogame.} \label{gta}
\end{figure}

Unlike previous works, our method does not require any human annotations. We automatically generate the annotations by defining the structure of the colon and polyps and transferring the location of the polyps to a 2D mask. The key difference between our approach and other state-of-the-art works is that we combine 3D rendering and generative networks. \hlA{Firstly}, the 3D engine defines the structure of the image and generates the annotations. \hlA{Secondly}, the adversarial network makes the images realistic. Note that similar unsupervised methods have also been successfully applied in other domains like crowd counting. For example, \citet{wang2019learning} render crowd images from a video game, as shown in Figure~\ref{gta} and then use a CycleGAN to increase the realism.

\subsection{Unpaired image translation}
Given two sets of images from two domains, unpaired image translation is used to transform images from one domain to the other. Ideally, the content of the images is preserved while their style is transformed to the new domain. Image translation models are divided in two main groups: two-sided and one-sided.

Two-sided translation models \citep{CycleGAN2017, Huang_2018_ECCV, Lee_2018_ECCV} are bijective, i.e. they transform data from the source domain to the target domain and vice versa thanks to a cycle consistency loss that helps both generative models to converge~\citep{CycleGAN2017}. These models are memory-intensive because they have to learn both translations. In general, they have at least four models (two generators and two discriminators), which also increases the training time.

\begin{figure}
\centering
\includegraphics[width=400px]{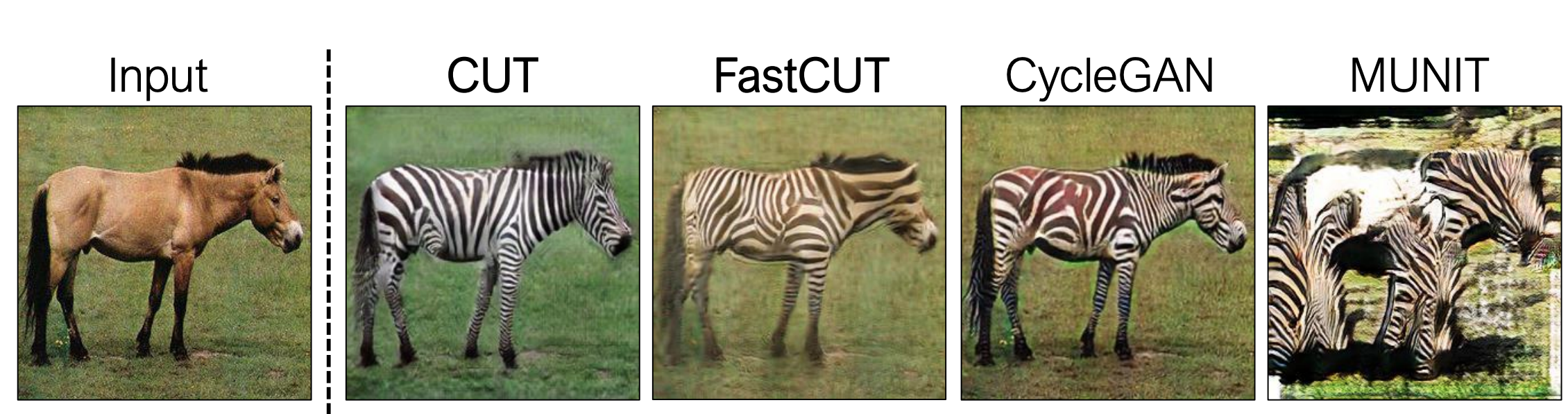}
\caption{Qualitative analysis of various approaches applied to the horse->zebra transformation.} \label{cut_vs_cyclegan}
\end{figure}

One-sided translation models \citep{park2020cut, Benaim2017OneSidedUD} only learn one transformation. As a consequence, they are lighter than the two-sided models. CUT \citep{park2020cut} extracts patches from the source image and learns the relationships between them in a self-supervised fashion. It uses a contrastive loss to maximize the mutual information between patches from the same region while minimizing the similarity between negative patches from the same image. This loss helps preserve the spatial context of the image. 
CUT \citep{park2020cut} is one of the faster and lighter unpaired image translation models because it is trained with patches from the same image, rather than from the rest of the dataset. In general, it offers better results than CycleGAN, as shown in Figure \ref{cut_vs_cyclegan}.

\section{Method}\label{sec3}
In this section we explain how we procedurally generate synthetic colon images and annotations using a 3D engine. Then, we describe the CycleGAN-based methodology we initially proposed in~\citep{2021_AICS_synthColon} to train a polyp segmentation model using synthetic data,  and CUT-seg, the improved single-stage approach.


\subsection{3D colon generation}

\begin{figure}[t]
\centering
\includegraphics[width=200px]{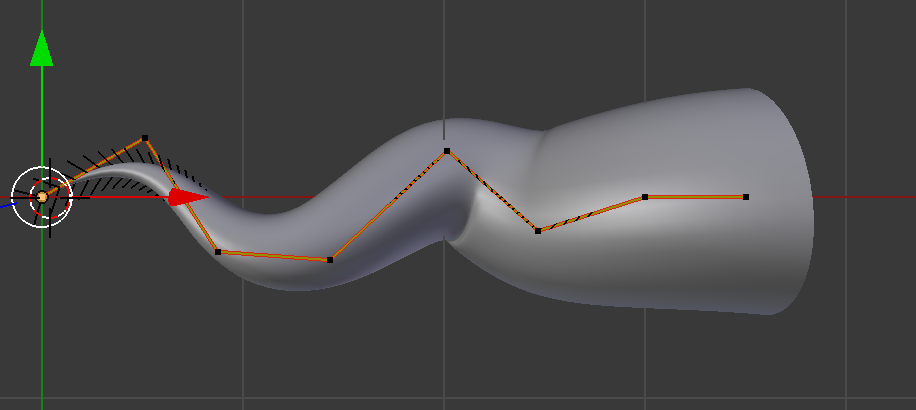}
\caption{The structure of the synthetic colon is composed of 7 segments to simulate the curvature of the intestinal tract.} \label{wireframe}
\end{figure}

The 3D colon and polyps are procedurally generated using Blender, a 3D engine that can be automated via scripting. Our 3D colon structure is a cone composed by 2454 faces. Vertices are randomly displaced following a uniform distribution in order to simulate the tissues in the colon. Additionally, the colon structure is modified by displacing 7 segments as in Figure~\ref{wireframe}.
For the textures we used a base color [0.80, 0.13, 0.18] (RGB). For each sample we shift the color to other tones by adding a 20\% of uniform noise to each channel.
One single polyp is used on every image, which is placed inside the colon. It can be either in the colon's walls or in the middle. Polyps are distorted spheres with 16384 faces. Samples with polyps occupying less than 2.6\% of the image are removed. This results in a dataset average polyp size of 5.87\%, which is within the values of the real datasets: the dataset with the smallest average polyp size is CVC-300 with 3.36\% and the largest is Kvasir with 16.46\%.

\begin{figure}[t]
\centering
\includegraphics[width=110px]{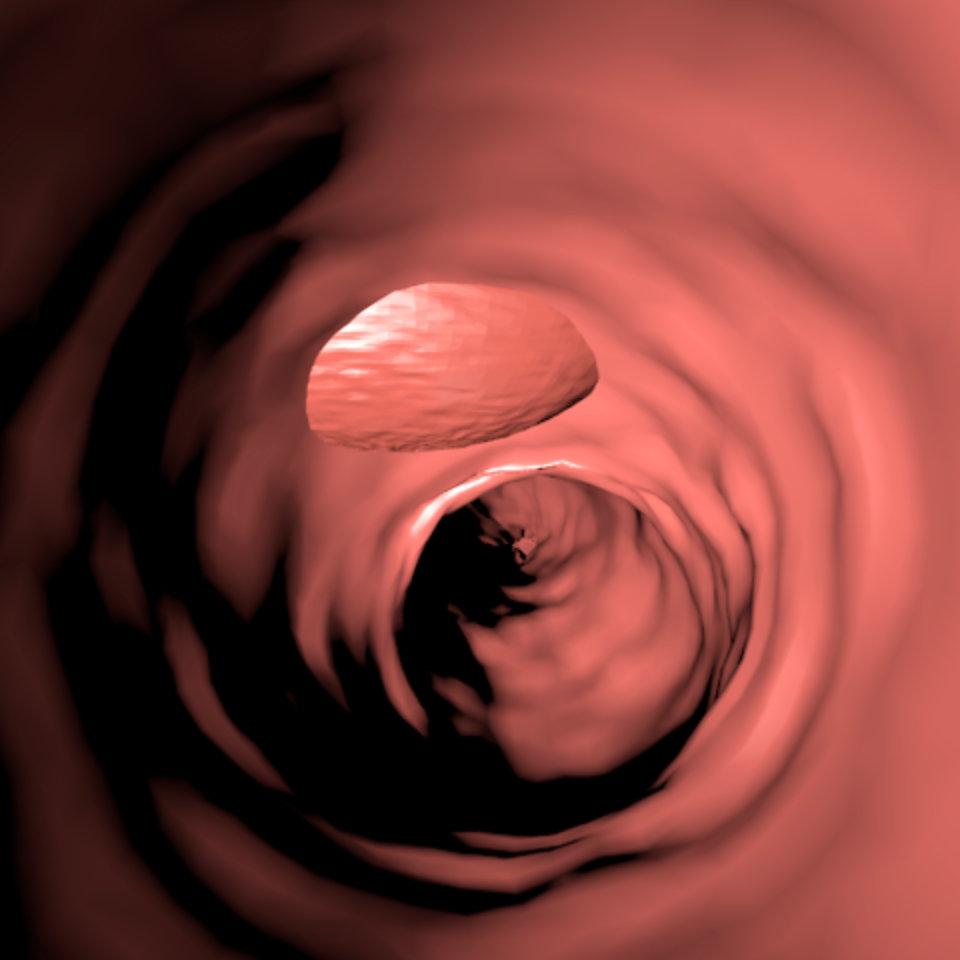}
\includegraphics[width=110px]{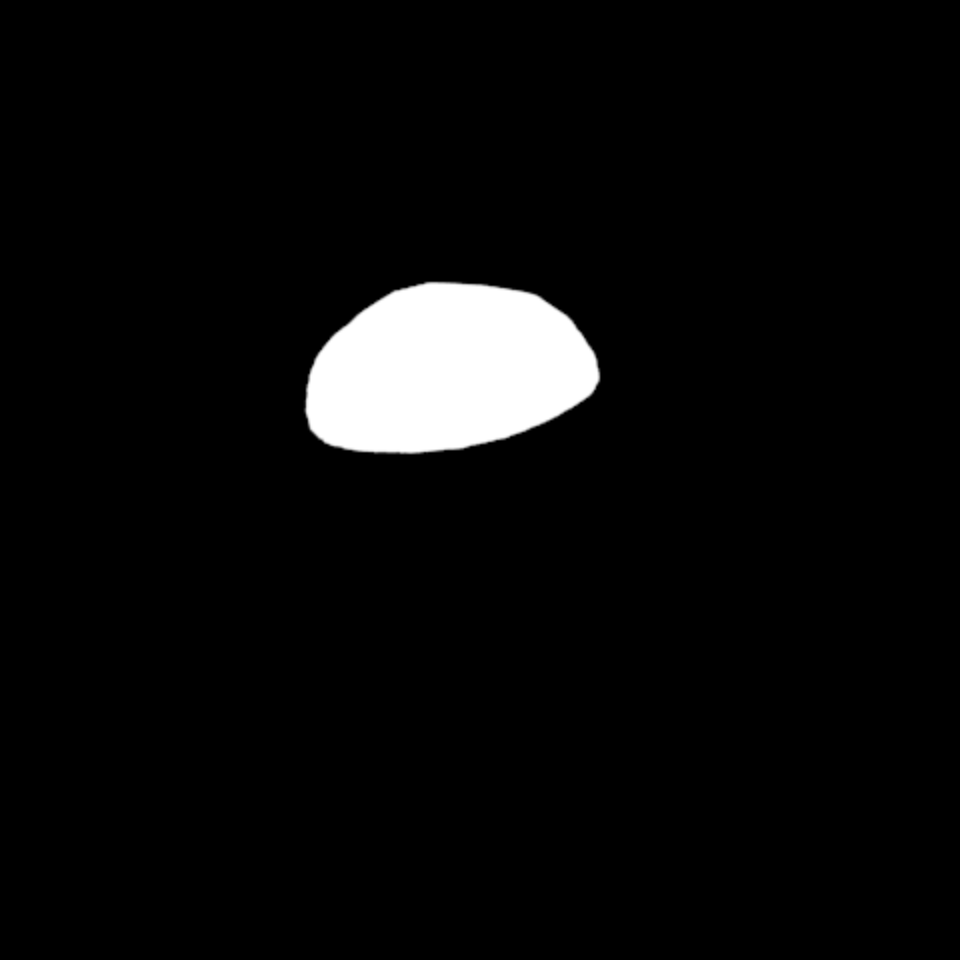}
\includegraphics[width=110px]{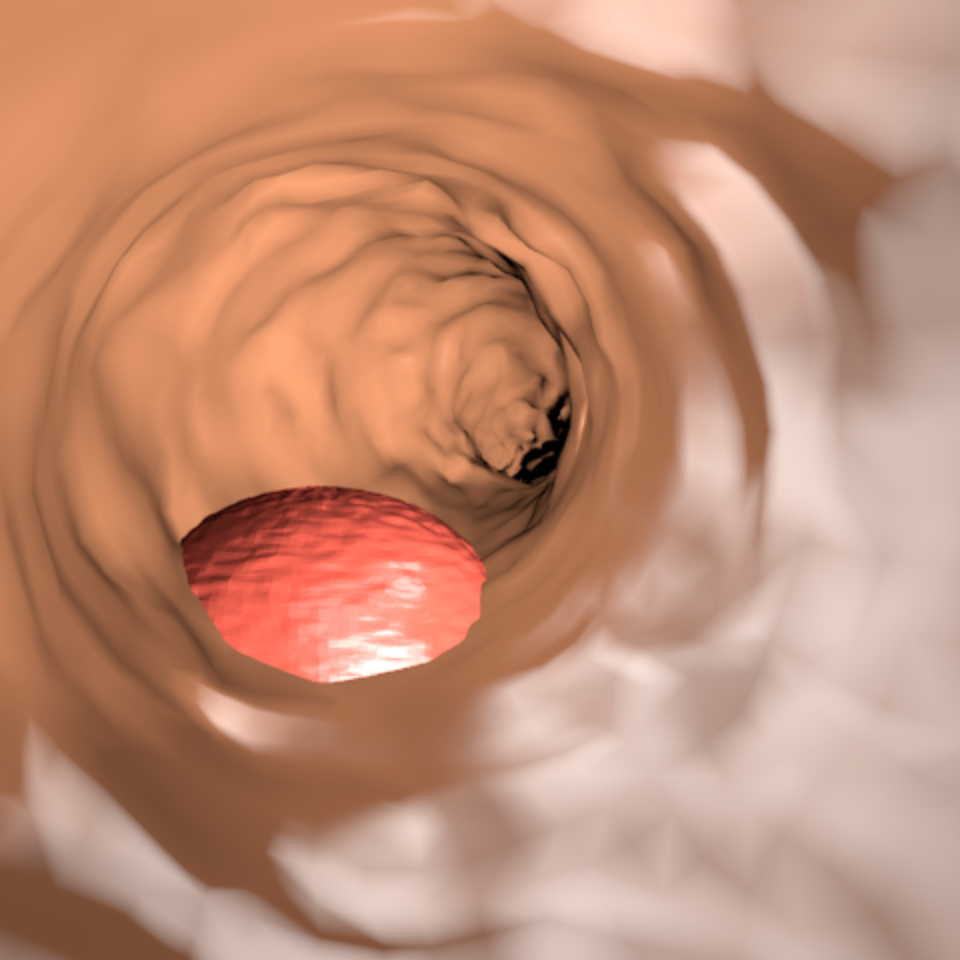}
\includegraphics[width=110px]{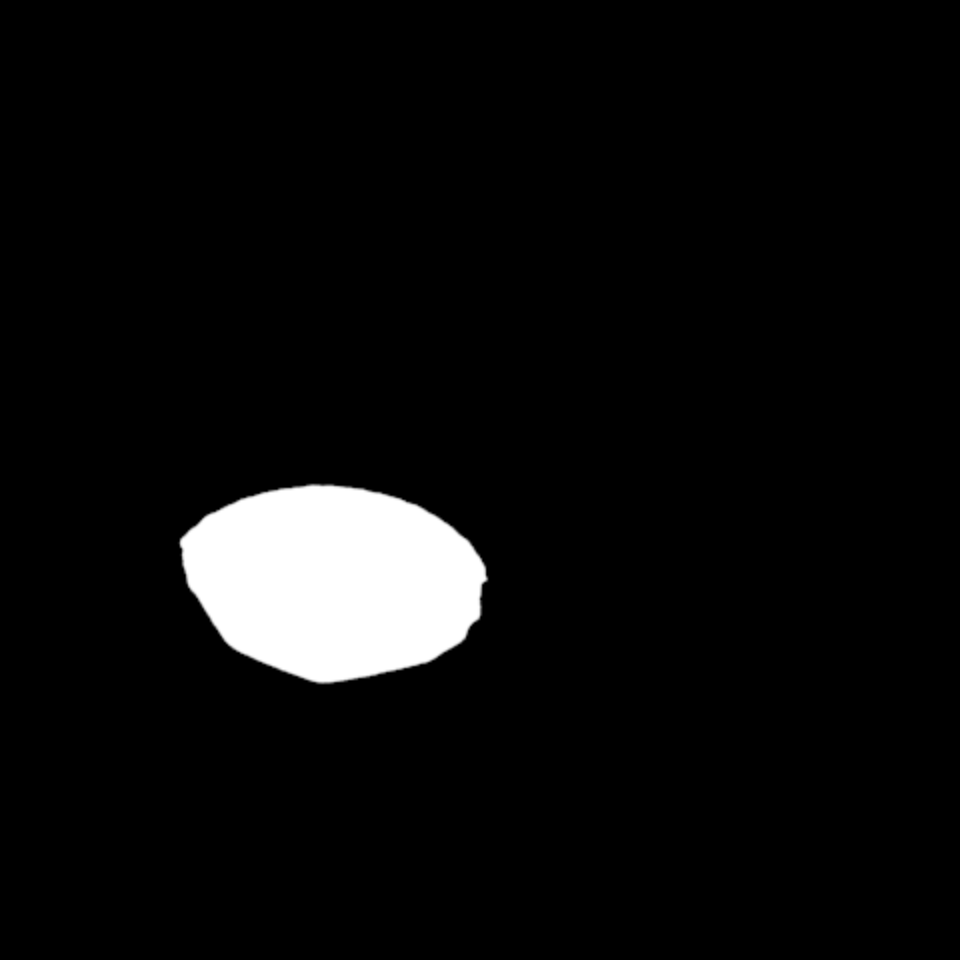}
\caption{Synthetic colons with corresponding annotations rendered using a 3D engine.} \label{synth_colons}
\end{figure}

Lighting is composed by a white ambient light, two white dynamic lights that project glare into the walls, and three negative lights that project black light at the end of the colon. We found that having a dark area at the end helps the generative models to understand the structure of the colon. The 3D scene must be similar to real colon images or the models will not properly translate the images to the real-world domain. Figure~\ref{synth_colons} illustrates the images and ground truth segmentation masks generated by the 3D engine.

\subsection{CycleGAN}
\begin{figure}
\centering
\includegraphics[width=400px]{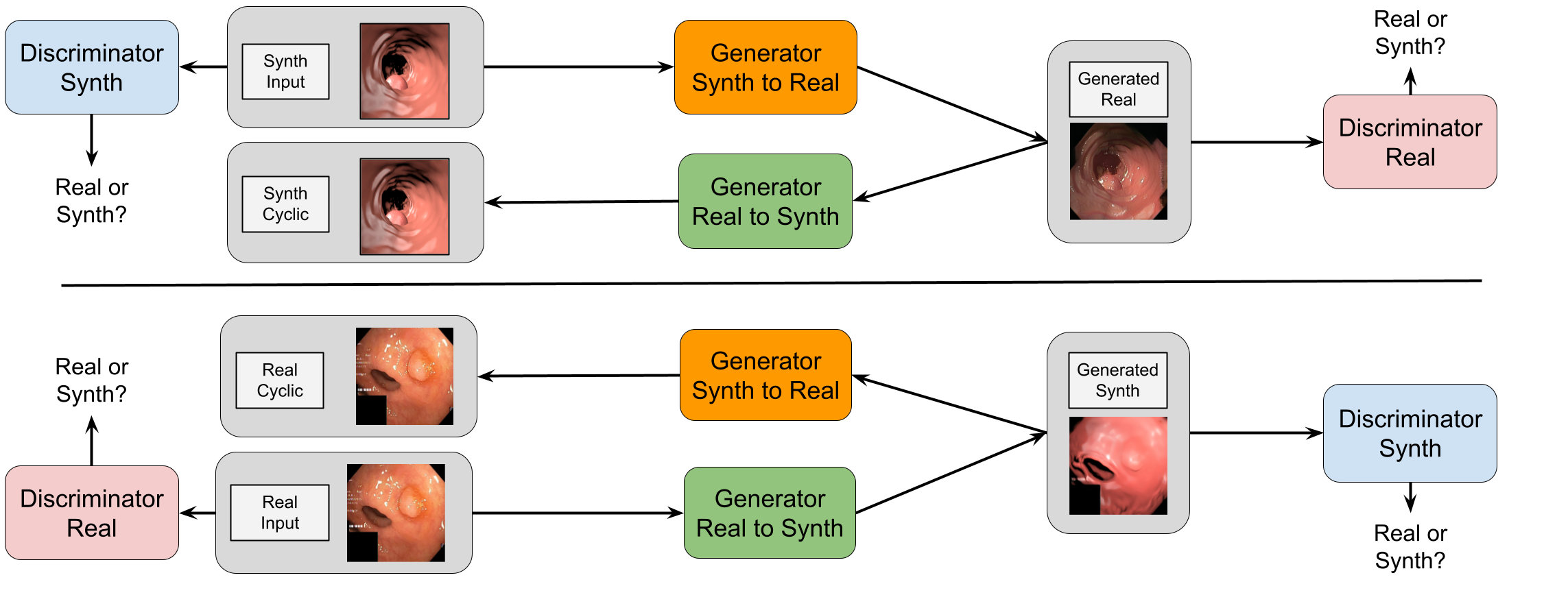}
\caption{Our CycleGAN-based architecture. We train two generator models that try to fool two discriminator models by changing the domain of the images.} \label{cyclegan_arch}
\end{figure}

\begin{figure}
\begin{tabular}{>{\centering\arraybackslash} m{3.85cm} >{\centering\arraybackslash} m{3.85cm} >{\centering\arraybackslash} m{3.85cm} >{\centering\arraybackslash} m{3.85cm}}

{Synthetic} &
\includegraphics[width=120px]{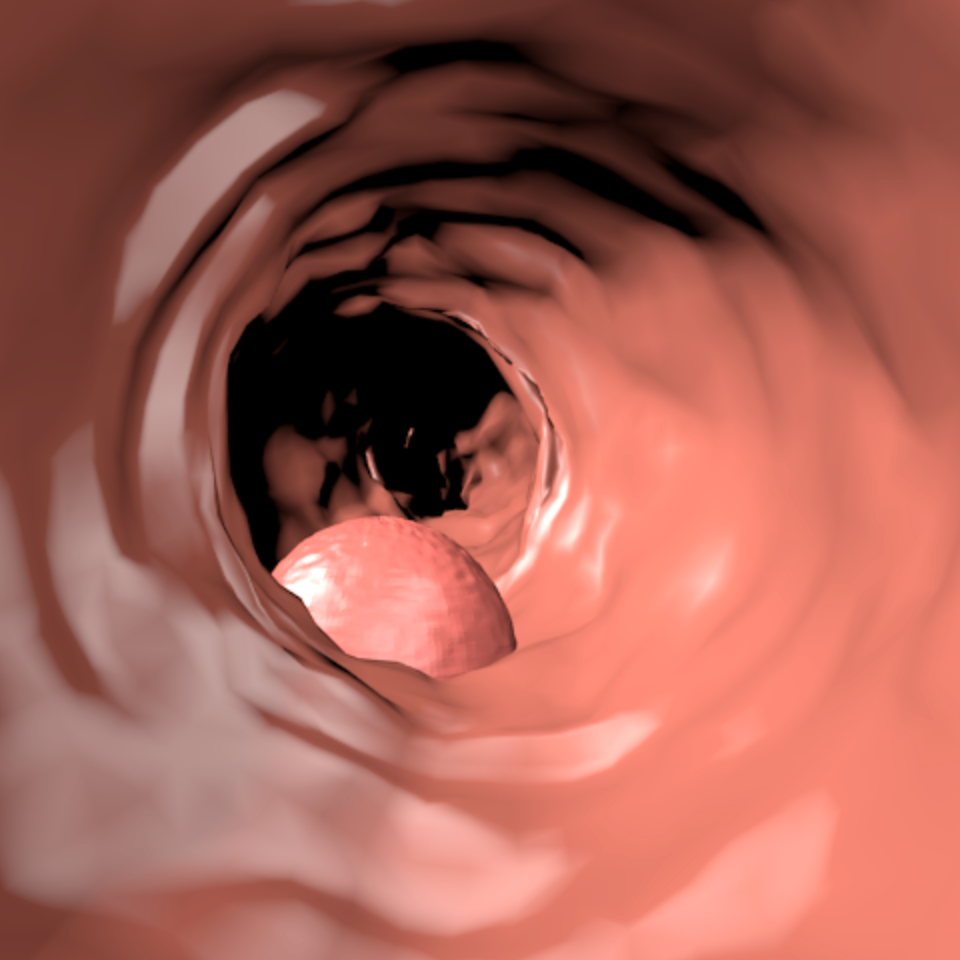} &
\includegraphics[width=120px]{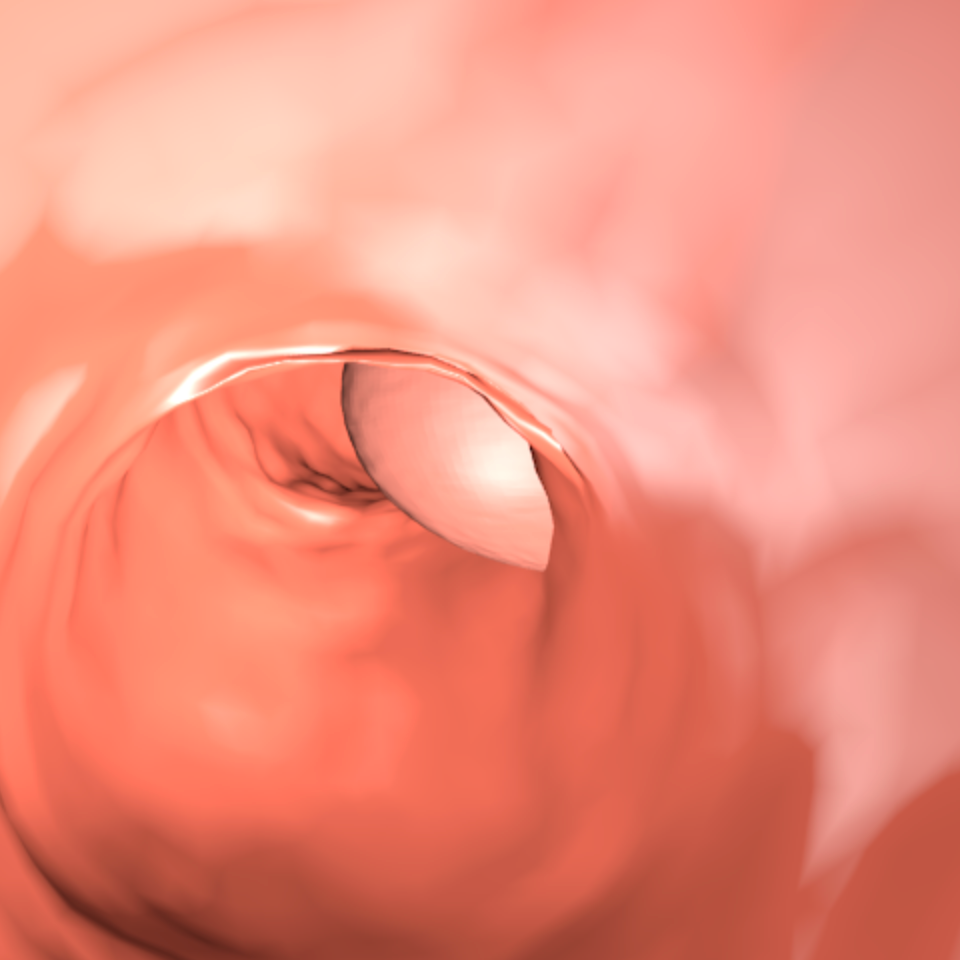} &
\includegraphics[width=120px]{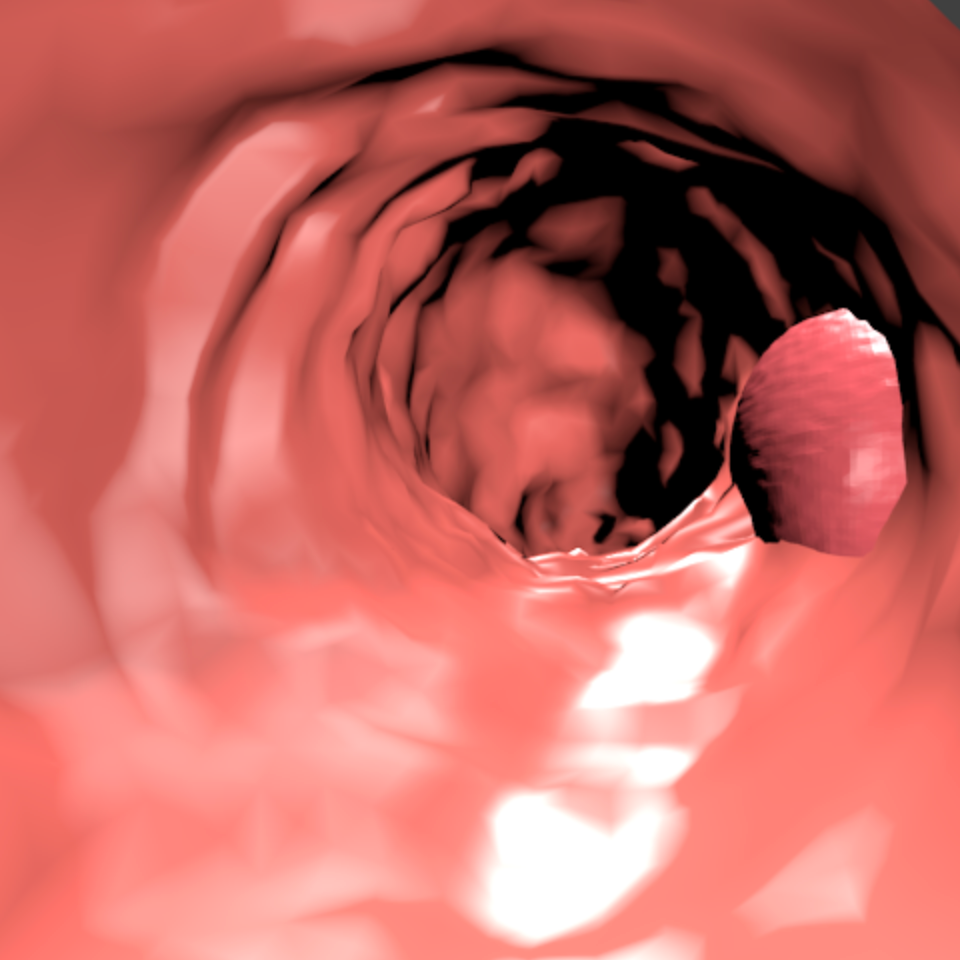}\\

{CycleGAN} &
\includegraphics[width=120px]{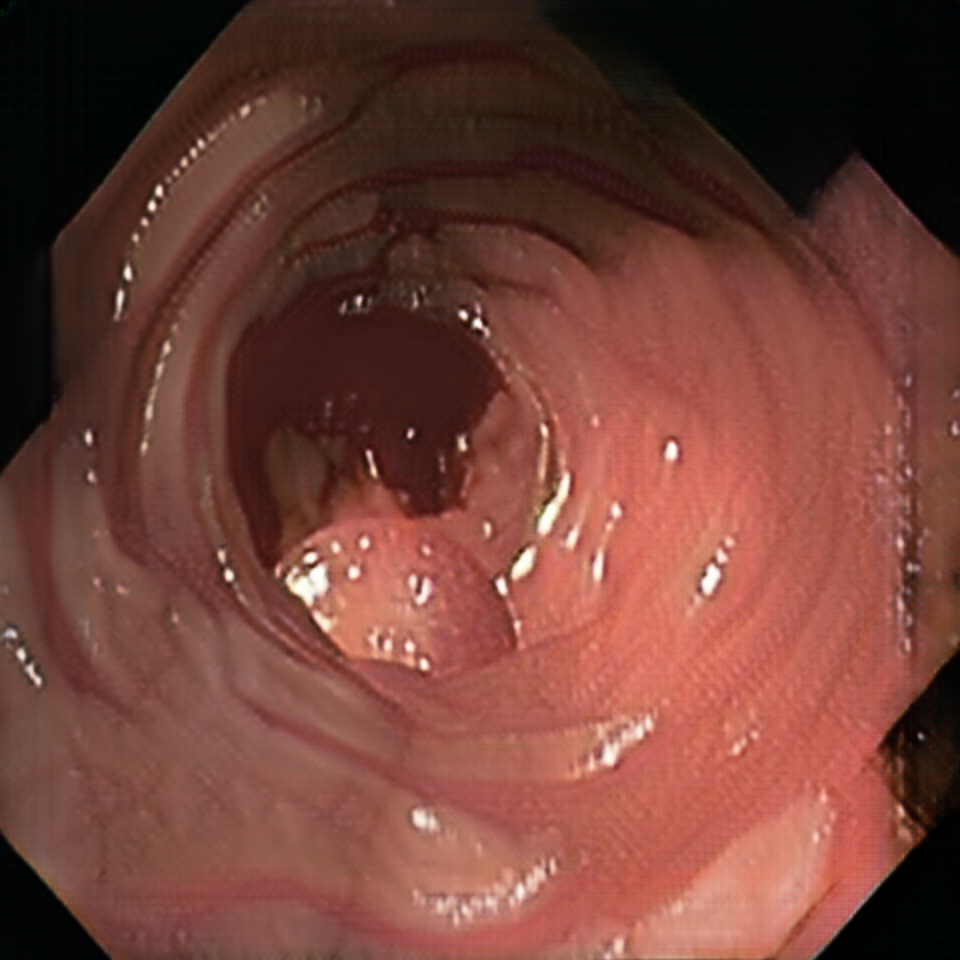} &
\includegraphics[width=120px]{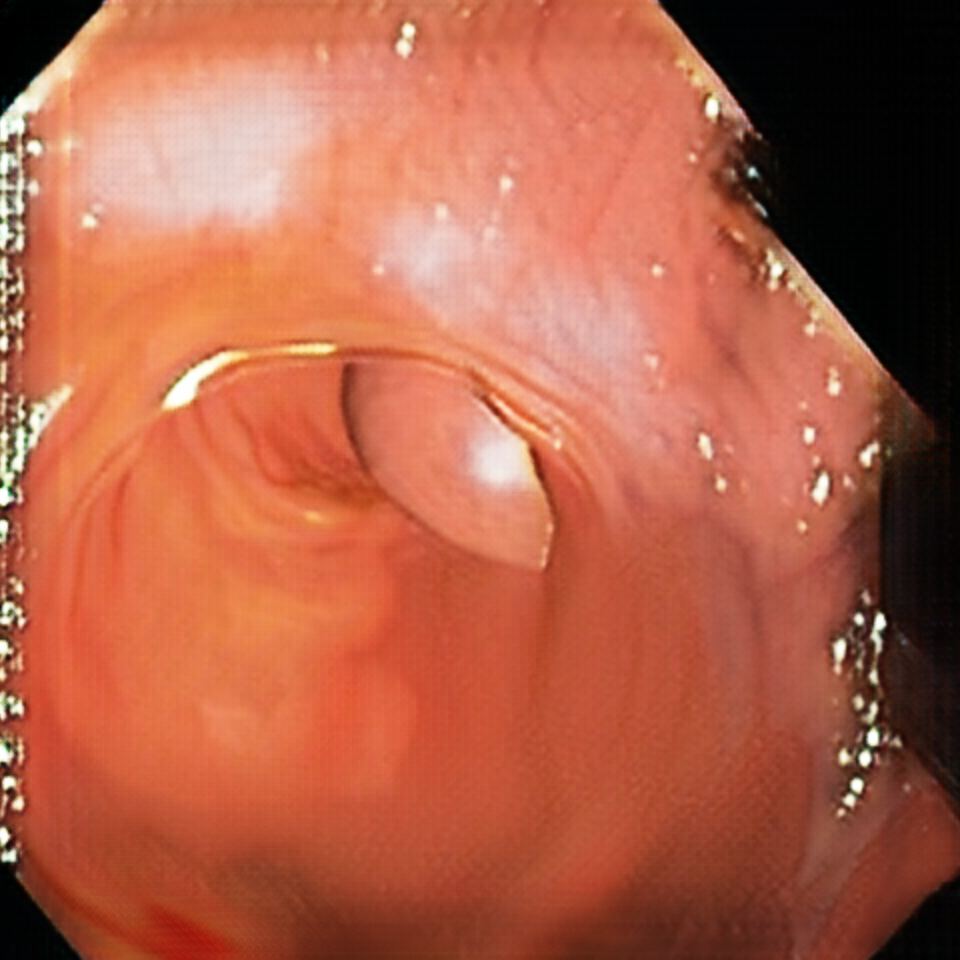} &
\includegraphics[width=120px]{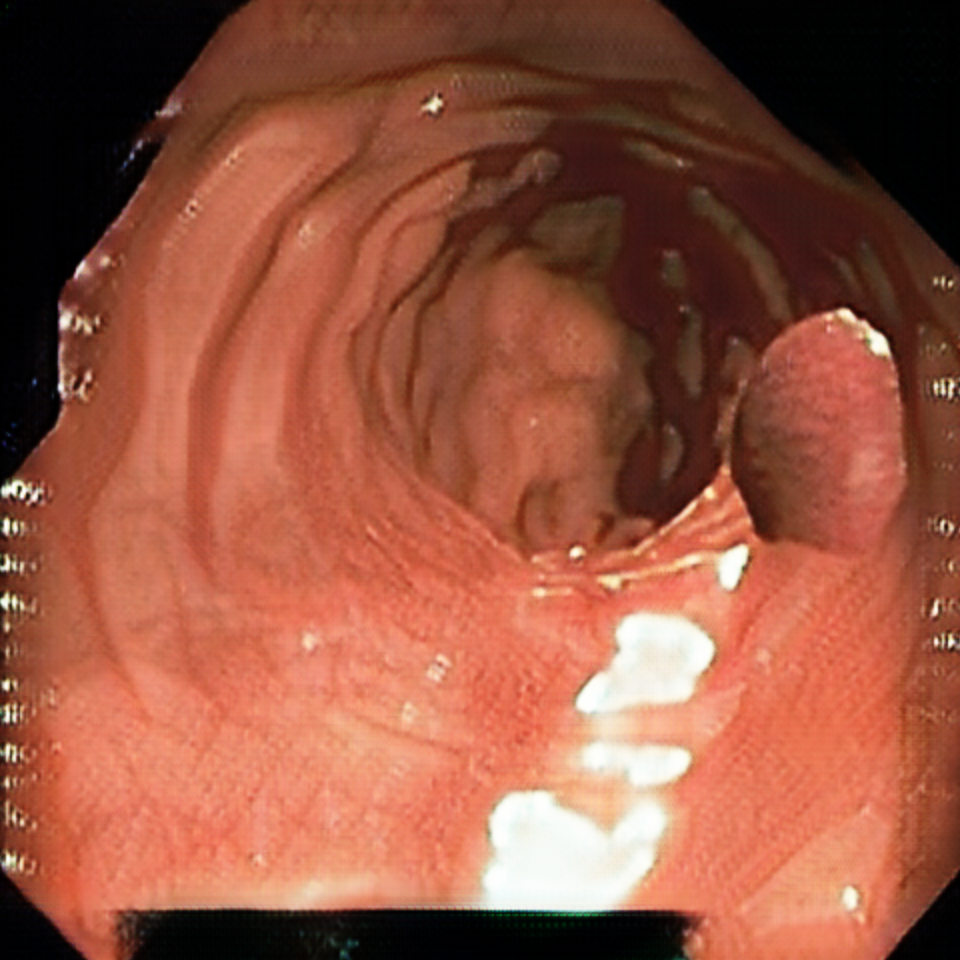}\\

{CUT} &
\includegraphics[width=120px]{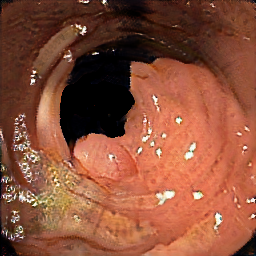} &
\includegraphics[width=120px]{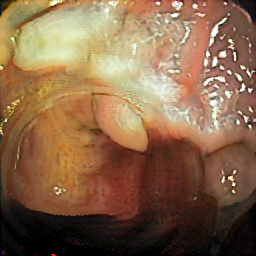} &
\includegraphics[width=120px]{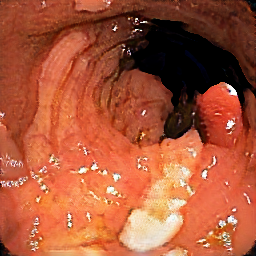} \\

{CUT-seg} &
\includegraphics[width=120px]{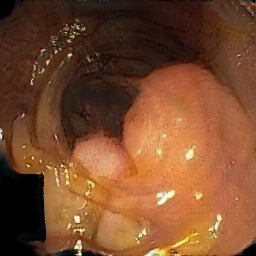} &
\includegraphics[width=120px]{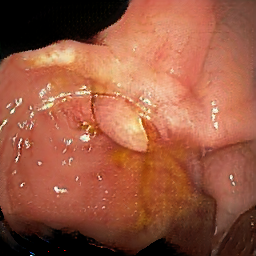} &
\includegraphics[width=120px]{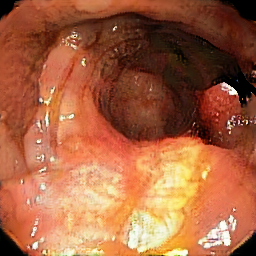}\\

{CUT-seg single image} &
\includegraphics[width=120px]{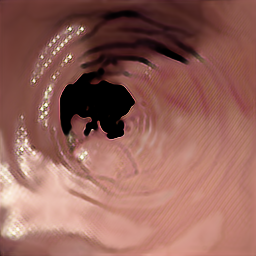} &
\includegraphics[width=120px]{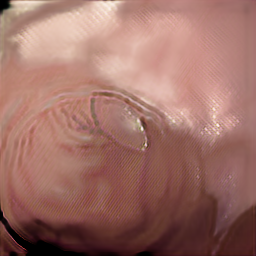} &
\includegraphics[width=120px]{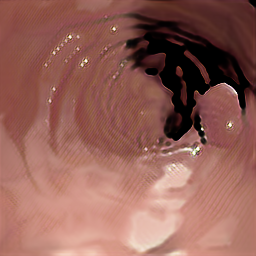}\\
\end{tabular}
\caption{Synthetic images (first row), CycleGAN generated images (second row), CUT generated images (third row), CUT-seg generated images (fourth row), and CUT-seg generated images with only one real sample as reference (fifth row).} \label{fake_samples}
\end{figure}

A standard CycleGAN, composed of two generators and two discriminators, is trained using real images from colonoscopies and synthetic images generated using the 3D engine as depicted in Figure~\ref{cyclegan_arch}. A generator-discriminator set translates synthetic images into real-looking images and the other set translates real images into synthetic-looking images. We train a CycleGAN for 200 epochs and then infer real images from synthetic ones with the corresponding generator (denoted as ``Generator Synth to Real'' in Figure~\ref{cyclegan_arch}), producing realistic colon images. Figure~\ref{fake_samples} displays synthetic images before (first row) and after (second row) the CycleGAN domain adaptation. Note that the position of the polyps is not altered. Hence, the ground truth information generated by the 3D engine is preserved.

The CycleGAN model is trained following an adversarial setup where the objective is defined by:
\begin{equation}
    \mathcal{L}_{\text{GAN}}(G, D_y, X, Y) +\mathcal{L}_{\text{GAN}}(F, D_x, Y, X) + \mathcal{L}_{\text{cyc}}(G, F),
\label{iou_eq}
\end{equation}
where $\mathit{G}$ is the real image generator, $\mathit{D_y}$ is the real discriminator, $\mathit{D_x}$ is the synthetic discriminator, $\mathit{F}$ is the synthetic images generator, $\mathit{X}$ are the synthetic domain samples, and $\mathit{Y}$ are the real domain samples. The term $\mathcal{L}_{\text{GAN}}(\mathit{G, D_y, X, Y})$ corresponds to the adversarial loss~\citep{2014_NeurIPS_gans} for the generator $\mathit{G}$, which maps from $\mathit{X}$ to $\mathit{Y}$, and its discriminator $\mathit{D_y}$, the term $\mathcal{L}_{\text{GAN}}(\mathit{F, D_x, Y, X})$ is the equivalent loss in the other direction to map from $\mathit{Y}$ to $\mathit{X}$, and $\mathcal{L}_{\text{cyc}}(\mathit{G, F})$ is the cycle consistency loss and enforces the translations of both generators to be cycle-consistent, 
i.e. the image generated by $\mathit{D}$ and its translation back to the original domain with $\mathit{F}$ should be similar:
\begin{equation}
    \mathcal{L}_{\text{cyc}}(G, F) = ||F(G(x)) - x||_1 + ||G(F(y)) - y||_1,
\label{cycle}
\end{equation}
where $\mathit{x}$ is an input image from $\mathit{X}$, the synthetic domain in our case, and $\mathit{y}$ an input image from $\mathit{Y}$, the real domain.

After creating a synthetic dataset that has been adapted to the real colon textures, we train an image segmentation model.
We used the HarDNeT-MSEG \citep{huang2021hardnetmseg} model architecture because of its real-time performance and high accuracy. We use the same hyperparameter configuration as in the original paper.

\subsection{CUT-seg}

CUT-seg attaches a HarDNeT-MSEG polyp segmentation architecture to a CUT generator~\citep{park2020cut}. The generative and segmentation models are jointly trained to generate realistic images with a polyp in a specific position matching the annotation mask. Combining both models is possible when GPU availability is limited because, unlike CycleGAN, CUT is more memory-efficient, leaving  considerable memory on the GPU for the segmentation model. Figure~\ref{fake_samples} shows examples of image translation using CUT alone (third row) and CUT-seg, jointly training CUT and the segmentation model (fourth row).



CUT builds on top CycleGAN and replaces the cycle consistency term with a contrastive term to encourage spatial consistency in the generated image. The resulting model does not need a secondary set of generator-discriminator models to regularize the training. The CUT loss function is given by:
\begin{equation}
    \mathcal{L}_{\text{GAN}}(G, D, X, Y) + \lambda_{X}\mathcal{L}_{\text{PatchNCE}}(G, H, X) + \lambda_{Y}\mathcal{L}_{\text{PatchNCE}}(G, H, Y),
\label{cut_loss}
\end{equation}
where $\mathcal{L}_{\text{PatchNCE}}(\mathit{G, H, X})$ is the contrastive term that encourages spatial consistency with the source image in $\mathit{X}$ (synthetic image in our case). This term encourages input-output patches from a particular location in an image to be close in the feature space, and far apart from other patches in the image. 
$\mathit{H}$ are the weights of a two-layer perceptron that projects the patches to the feature space, and $\mathit{\lambda_{X}}$ and $\mathit{\lambda_{Y}}$ are hyperparamters that control the contribution of the corresponding contrastive terms.

Our CUT-seg method includes a segmentation loss that is optimized alongside the generator. In particular, we substitute the $\mathcal{L}_{\text{GAN}}(G, D, X, Y)$ term in Eq.~\eqref{cut_loss} with: %
\begin{equation}
    \mathcal{L}_{\text{GAN}}(G, D, X, Y) + \lambda_{S}\mathcal{L}_{\text{Seg}}(S),
\label{cut-seg_segmentation_loss}
\end{equation}
where $\mathcal{L}_{Seg}(\mathit{S})$ is the mean Dice loss on the segmentation masks inferred from the images generated by $\mathit{D}$, %
$\mathit{S}$ is the segmentation model, and $\lambda_{\mathit{S}}$ a hyperparameter that controls the weight of the segmentation term.

\section{Synth-Colon}\label{sec4}
We publicly release Synth-Colon, a synthetic dataset for polyp segmentation. It is the first dataset generated using zero annotations from medical professionals. The dataset is composed of \numprint{20000} images with a resolution of 500$\times$500.  Synth-Colon additionally includes realistic colon images generated with our CycleGAN and the Kvasir training set images. Additionally, Synth-Colon can also be used for the colon depth estimation task~\citep{rau2019implicit} because we provide depth and 3D information for each image. It helps doctors to verify that all the surfaces in the colon have been analyzed. Figure~\ref{dataset} shows some examples from the dataset.
In summary, Synth-Colon includes:
\begin{itemize}

\item Synthetic images of the colon and one polyp.
\item Masks indicating the location of the polyp.
\item Realistic images of the colon and polyps generated using our CycleGAN baseline and the Kvasir dataset.
\item Depth images of the colon and polyp.
\item 3D meshes of the colon and polyp in OBJ format.

\end{itemize}

\section{Experiments}\label{sec5}

Experiments are evaluated using the mean Dice (mDICE) and mean intersection over union (mIoU). The mDICE is defined by:
\begin{equation}
    \mathrm{mDice} = \frac{2 \times tp}{2\times tp+fp+fn},
\end{equation}
and the mIoU is given by:
\begin{equation}
    \mathrm{mIoU} = \frac{tp}{tp+fp+fn},
\label{iou}
\end{equation}
where in both forumlae, $\mathit{tp}$ is the number of true positives, $\mathit{fp}$ the number of false positives, and $\mathit{fn}$ the number of false negatives.

We used a RTX 2080Ti GPU with 11GB of memory to run the experiments. We found the best results by setting the learning rate of CUT-seg to $10^{-5}$, the weight of the segmentation term  $\mathit{\lambda_{S}}$ to $10^{-3}$, and the two weights of the contrastive terms $\mathit{\lambda_{X}}$ and $\mathit{\lambda_{Y}}$ to $0.5$ as in the original CUT paper~\citep{park2020cut}. This value is used by both generator and discriminator optimizers. Note that the segmentation model is optimized alongside the generator.

\subsection{Transductive evaluation}

Table~\ref{transductive} shows the results obtained with our CycleGAN-based baseline and CUT-seg model. Neither the CycleGAN-based model nor the CUT-seg model use any real human annotation, unlike the other approaches that are compared in the table. We evaluate our approach on five real polyp segmentation datasets in a transductive setup. This is a common setup in zero-shot learning~\citep{2015_TPAMI_transductive, 2015_ICCV_transductive, 2018_CVPR_transductive, 2019_NeurIPS_transductive} that explores the performance of a model when the unlabeled target data, the test set in our case, is available during training. Unlike the zero-shot setup, in our case the source data is also unlabeled, except for the synthetic images that come with free annotations. Transductive evaluation is a valuable setup to bypass the domain gap between target and source data, and better understand the performance of the algorithm in a specific target domain. In our case this addresses an inherent challenge in the training dataset: the samples from all the datasets are mixed in a single training set. Note that in inductive evaluation, only the labeled source data is available.

CUT-seg displays not only superior performance compared with our CycleGAN baseline, but also trains approximately 4 times faster. Results are satisfactory considering the fact that our synthetic annotations have been generated automatically. We found that training CUT-seg with only the images from the target dataset performs better than training it with all the datasets combined, indicating a domain gap among the real-world datasets.
In this setup we observe that CUT-seg outperforms the CycleGAN baseline in most of the datasets despite requiring less computation. These results are still below the fully supervised state-of-the-art HarDNet-MSEG~\citep{huang2021hardnetmseg} model but considerably reduce the performance gap between methods that use manual annotations and methods that do not.

\begin{table}
\centering
\caption{Evaluation of our synthetic approach on real-world datasets. The metrics used are mean Dice similarity index (mDice) and mean intersection over union (mIoU). Best results are highlighted in \ctB{blue} and the best results without human supervision are \underline{underlined}.}\label{transductive}
\addtolength{\tabcolsep}{2pt}    

\resizebox{\textwidth}{!}{%
\begin{tabular}{l@{\qquad}cccccccccccc}
  \toprule
  &
  \multicolumn{2}{c}{CVC-T} & \multicolumn{2}{c}{ColonDB} & \multicolumn{2}{c}{ClinicDB} & \multicolumn{2}{c}{ETIS} & \multicolumn{2}{c}{Kvasir} \\
  & mDice & mIoU & mDice & mIoU & mDice & mIoU & mDice & mIoU & mDice & mIoU \\
  \midrule
  U-Net~\citep{ronneberger2015u} & 0.710 & 0.627 & 0.512 & 0.444 & 0.823 & 0.755 & 0.398 & 0.335 & 0.818 & 0.746 \\
  SFA~\citep{fang2019selective} & 0.467 & 0.329 & 0.469 & 0.347 & 0.700 & 0.607 & 0.297 & 0.217 & 0.723 & 0.611 \\
  PraNet~\citep{fan2020pranet} & 0.871 & 0.797 & 0.709 & 0.640 & 0.899 & 0.849 & 0.628 & 0.567 & 0.898 & 0.840 \\
  HarDNet-MSEG~\citep{huang2021hardnetmseg} & \ctB{0.887} & \ctB{0.821} & \ctB{0.731} & \ctB{0.660} & \ctB{0.932} & \ctB{0.882} & \ctB{0.677} & \ctB{0.613} & \ctB{0.912} & \ctB{0.857} \\
  \midrule
  CycleGAN-based & \underline{0.703} & \underline{0.635} & 0.521 & \underline{0.452} & 0.551 & 0.475 & 0.257 & 0.214 & \underline{0.759} & 0.527 \\
  CUT-seg & 0.700 & 0.613 & \underline{0.546} & 0.396 & \underline{0.719} & \underline{0.573} & \underline{0.540} & \underline{0.384} & 0.702 & \underline{0.621} \\
  \bottomrule
\end{tabular}}
\end{table}


\subsection{Synthetic dataset size}
\begin{figure}
\centering
\caption{Evaluation of CUT-seg with varying amounts of synthetic data on the Kvasir dataset. The performance is measured using the mDice metric, each experiment reports the average mDice across three runs and the error bars indicate the standard deviation.}\label{synth_size}
\includegraphics[width=0.8\textwidth]{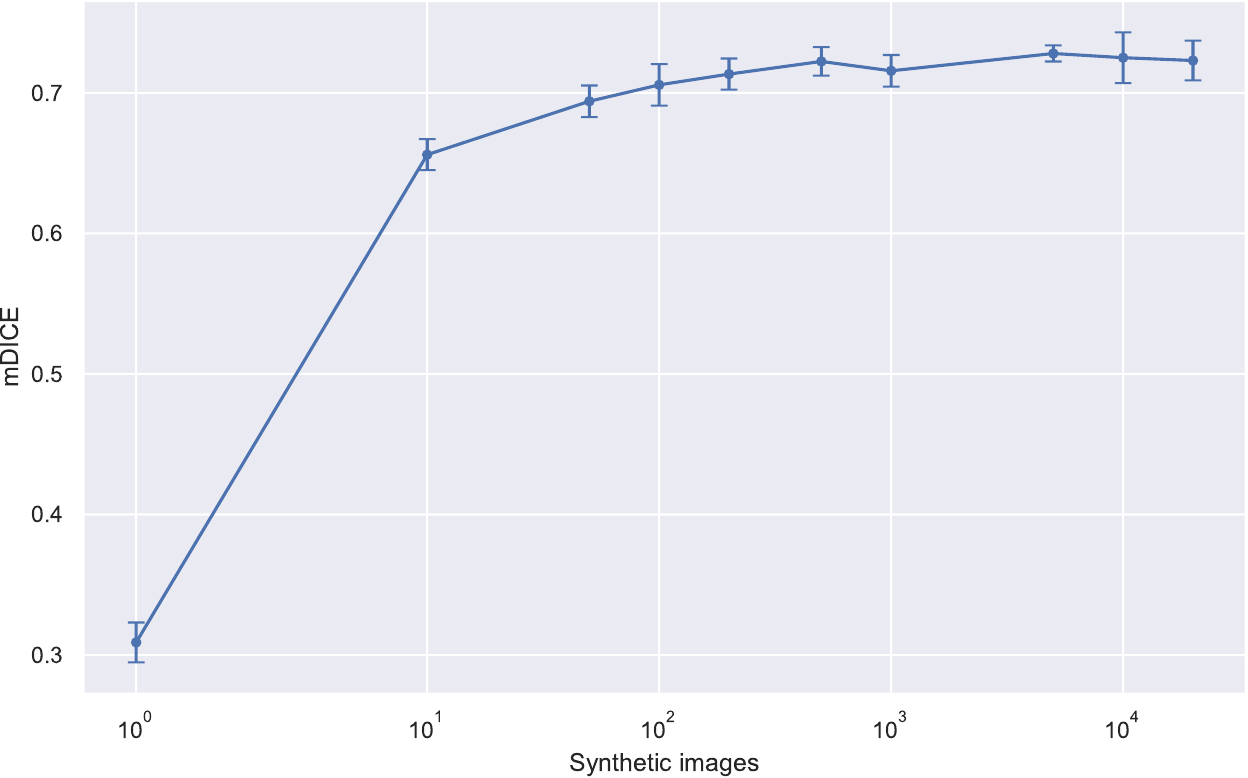}
\end{figure}

We explore how many synthetic images are needed to successfully train CUT-seg. For this experiment we train our model using 100 real images from the Kvasir dataset and a varying number of synthetic images.
Figure~\ref{synth_size} demonstrates that CUT-seg benefits from a large number of synthetic samples. While the best results are obtained with the largest amounts of samples, CUT-seg reaches near peak performance when training with 100 or more synthetic images.


\subsection{Single reference image}
We found that using only a single real image (without ground truth) is sufficient to successfully translate images from the synthetic domain to the real world as shown in Figure~\ref{fake_samples} (fifth row). Table \ref{single_image} shows that the performance improves in some datasets when using only a single real image instead of the full real dataset. 

We hypothesize that the reason for this phenomenon is that real datasets contain images that are more representative than others. Training with all images in the dataset will inevitably use samples that do not characterize the dataset. When training using a single representative image as a reference the results are sometimes better because the generated images are more similar to the test set.

\begin{table}
\centering
\caption{Comparison between training CUT-seg using the full dataset or a single real image. The metrics used are mean Dice similarity index (mDice) and mean Intersection over Union (mIoU). Best results are highlighted in \textbf{bold}.}\label{single_image}
\addtolength{\tabcolsep}{2pt}    

\resizebox{\textwidth}{!}{%
\begin{tabular}{l@{\qquad}cccccccccccc}
  \toprule
  &
  \multicolumn{2}{c}{CVC-T} & \multicolumn{2}{c}{ColonDB} & \multicolumn{2}{c}{ClinicDB} & \multicolumn{2}{c}{ETIS} & \multicolumn{2}{c}{Kvasir} \\
  & mDice & mIoU & mDice & mIoU & mDice & mIoU & mDice & mIoU & mDice & mIoU \\
  \midrule
    All real images & 0.700 & 0.613 & 0.546 & 0.396 & \textbf{0.719} & \textbf{0.573} & \textbf{0.540} & \textbf{0.384} & 0.702 & 0.621 \\
    One real image & \textbf{0.754} & \textbf{0.617} & \textbf{0.569} & \textbf{0.422} & 0.636 & 0.563 & 0.412 & 0.334 & \textbf{0.732} & \textbf{0.640} \\
  \bottomrule
\end{tabular}}
\end{table}

It is worth mentioning that the results depend strongly on which real image is used as reference and the initialization of the model. We trained $\mathit{N}=10$ runs on the Kvasir dataset with different reference images and measured a mean of 0.70 mDice and a standard deviation of 0.02. When training with all the images, the mean is 0.71 and the standard deviation is 0.01.
Future work will explore why some images improve the performance, and how to identify these images.

\section{Conclusions \& future work}\label{sec6}

We successfully trained a polyp segmentation model without human annotations by exploiting synthetically generated images: we used 3D rendering to generate the structure of the colon and generative adversarial networks to make the images look more realistic. We demonstrated that segmentation models trained on this data yield competitive results in several datasets, even outperforming fully supervised methods in some cases.
Furthermore, we propose an end-to-end model that jointly learns to generate realistic images and segment polyps, and \hlA{demonstrates} that a joint training allows for a faster learning and provides better results than the two-stage counterpart,~i.e. training a generative model and segmentation model separately.
With this study, we bring a step closer the application of synthetic data to the medical domain and hope that this research motivates further exploration of how to \hlA{align} these two domains in the future. As future work, we will explore strategies to better leverage larger synthetic datasets and the applicability of our method in the video domain.

\bibliography{wileyNJD-APA}%

\section*{Author Biography}

\begin{biography}{\includegraphics[width=60pt,height=70pt]{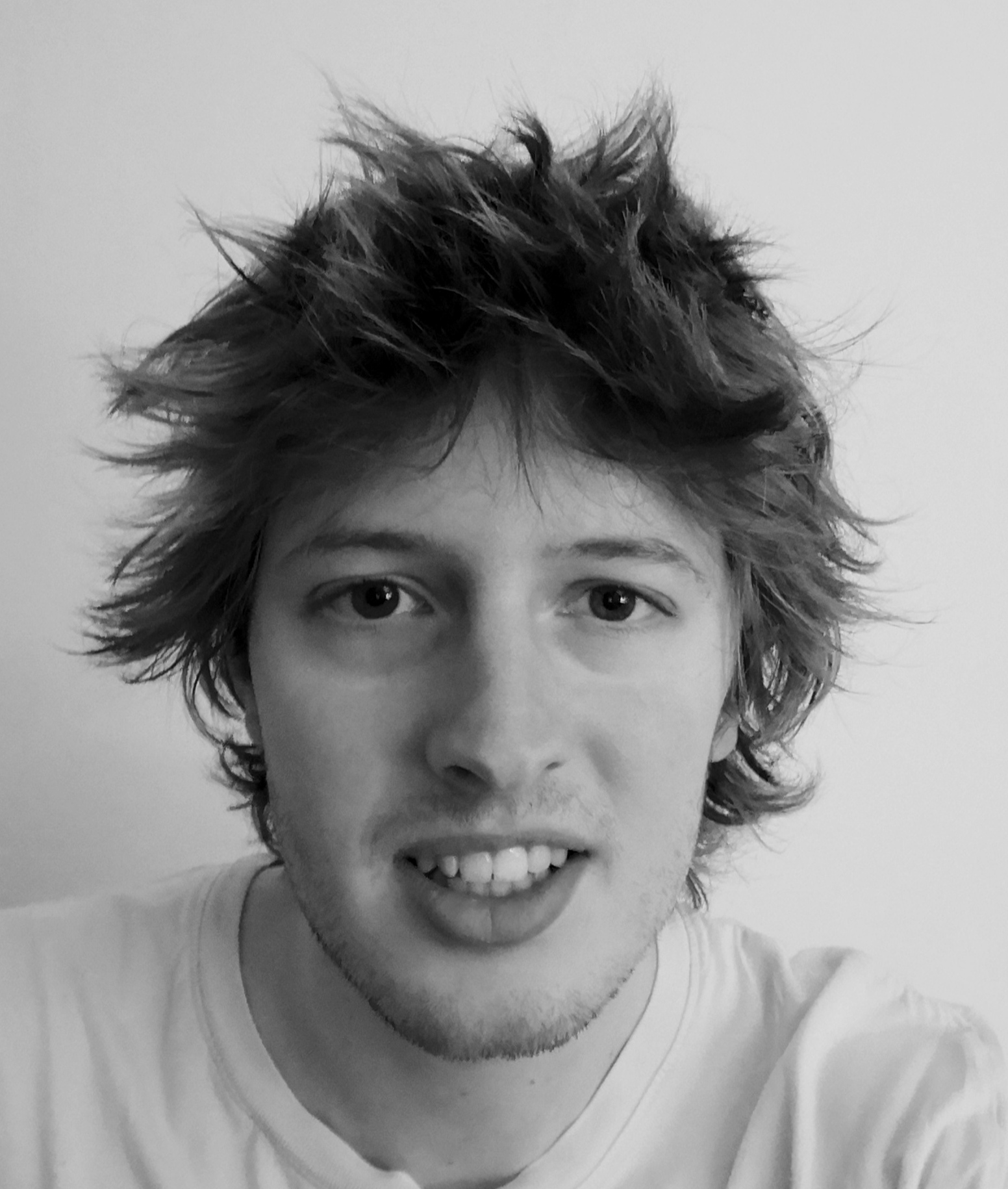}}{\textbf{Enric Moreu.} Enric Moreu is a PhD candidate in the School of Electronic Engineering at Dublin City University and a researcher in the Insight SFI Research Centre for Data Analytics. He has a BSc in Telecommunications Engineering (2017) from Universitat Politècnica de Catalunya. His research in synthetic data for computer vision is funded by the Marie Skłodowska-Curie Actions.
\vspace{1.3cm}
}
\end{biography}

\begin{biography}{\includegraphics[width=60pt,height=70pt]{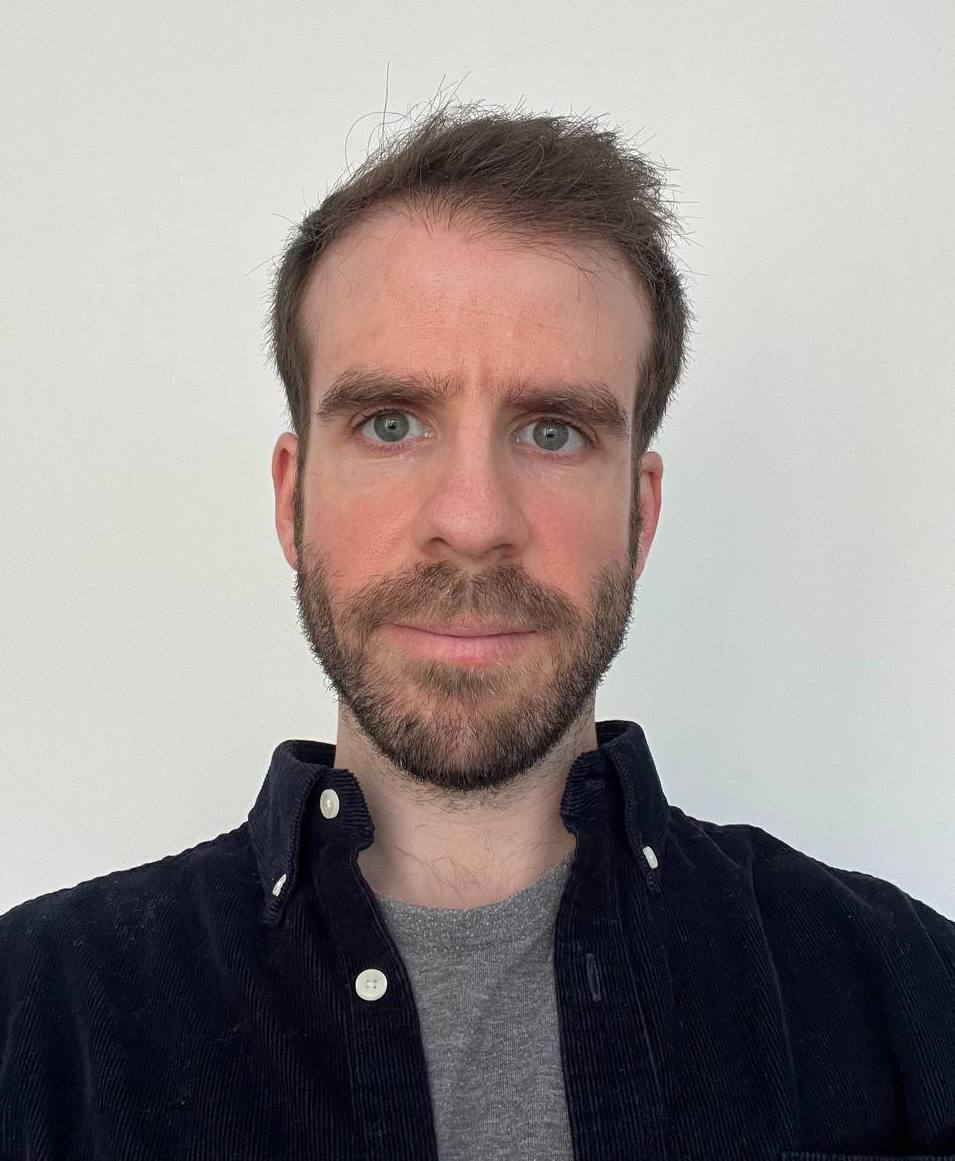}}{\textbf{Eric Arazo.} Dr Eric Arazo is a post-doctoral researcher in the School of Electronic Engineering at Dublin City University and a researcher in the Insight SFI Research Centre for Data Analytics. He has a BEng and a MEng in Telecommunications Engineering (2015 and 2017) from Universitat Politècnica de Catalunya. He finished his BEng and MEng thesis work at Tallinn University of Technology and Dublin City University. During his research as a PhD candidate in Dublin City University, his primary research interests focused on the application of machine learning, and deep learning algorithms in weakly supervised setups for computer vision.}
\end{biography}

\begin{biography}{\includegraphics[width=60pt,height=70pt]{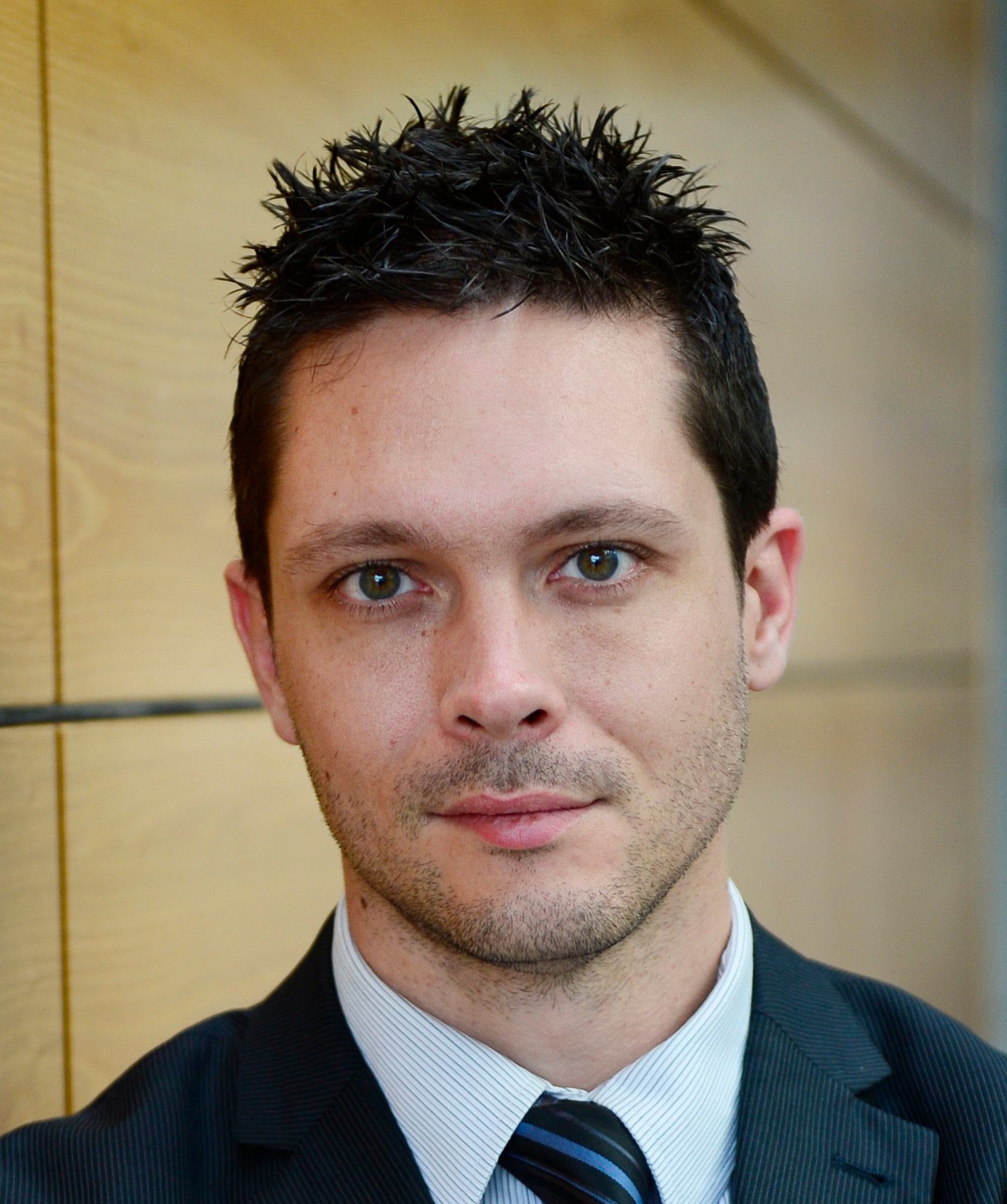}}{\textbf{Kevin McGuinness.} Dr Kevin McGuinness is an Assistant Professor in the School of Electronic Engineering at Dublin City University and SFI Funded Investigator in the Insight SFI Research Centre for Data Analytics.  He has a BSc (Hons) in Computer Applications (2005) and a PhD in Computer Vision (2009) from Dublin City University. Since 2009 he has been a postdoctoral researcher at the CLARITY Centre for Sensor Web Technologies, a Research Fellow at the Insight Centre for Data Analytics, and now teaches graduate-level data analysis and machine learning for the School of Electronic Engineering. Kevin has 100+ peer-reviewed publications focused on topics in computer vision, machine learning, and deep learning.}
\end{biography}

\begin{biography}{
\includegraphics[width=60pt,height=70pt]{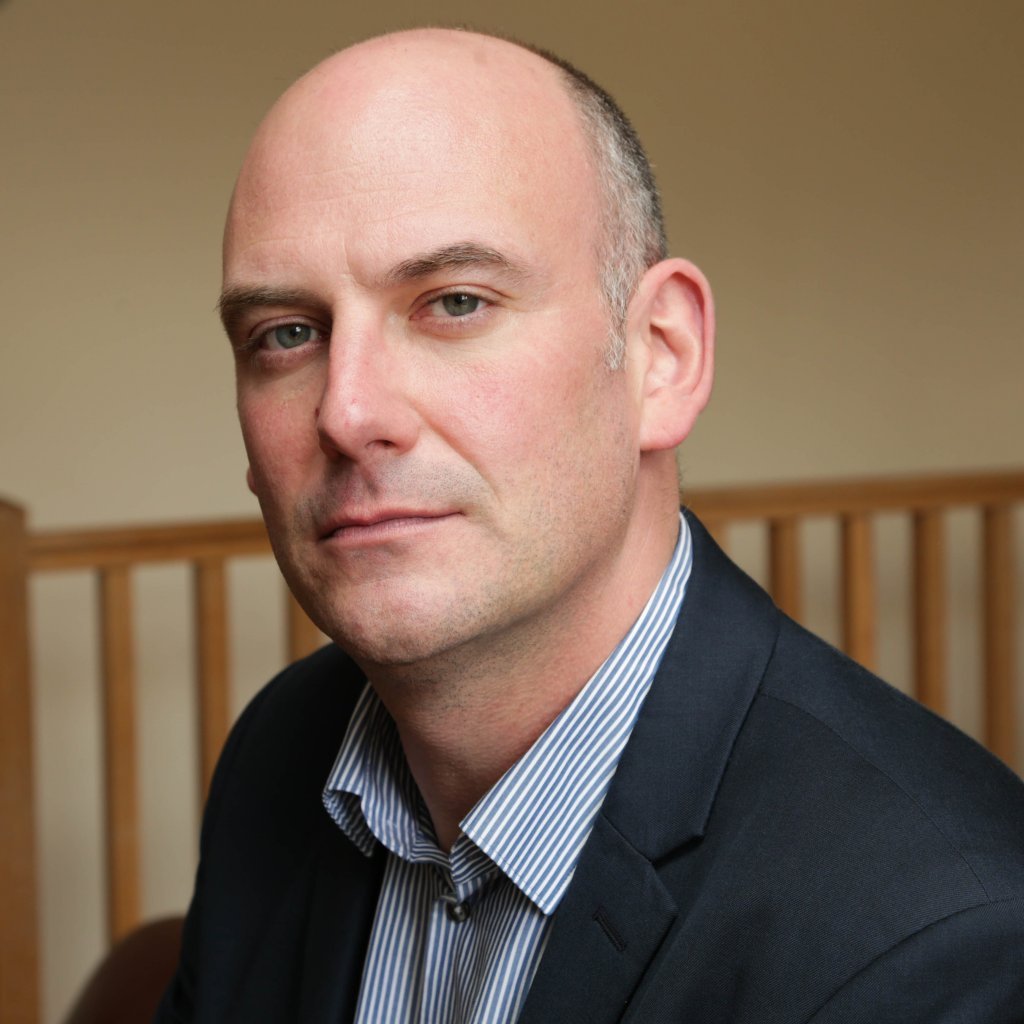}}{\textbf{Noel E. O'Connor.} Prof. Noel E. O’Connor is a Full Professor in the School of Electronic Engineering at Dublin City University (DCU) Ireland. He is CEO of the Insight SFI Research Centre for Data Analytics, Ireland’s largest SFI-funded research centre.  The focus of his research is in multimedia content analysis, computer vision, machine learning, information fusion and multi-modal analysis for applications in security/safety, autonomous vehicles, medical imaging, IoT and smart cities, multimedia content-based retrieval, and environmental monitoring. Since 1999 he has published over 400 peer-reviewed publications, made 11 standards submissions, and filed 7 patents. He is an Area Editor for Signal Processing: Image Communication (Elsevier) and an Associate Editor for ACM Transactions on Multimedia Computing, Communications, and Applications. He is a member of the ACM and IEEE.}
\end{biography}

\end{document}